\documentclass{article}

\usepackage{arxiv}

\usepackage[utf8]{inputenc} 
\usepackage[T1]{fontenc}    
\usepackage{hyperref}       
\usepackage{url}            
\usepackage{booktabs}       
\usepackage{amsfonts}       
\usepackage{nicefrac}       
\usepackage{microtype}      
\usepackage{lipsum}
\usepackage{graphicx}
\graphicspath{ {./images/} }

\usepackage{doi}
\usepackage{amsmath}
\usepackage{amssymb}
\usepackage{booktabs}
\usepackage{array}
\usepackage{multirow}
\usepackage{amsmath}
\usepackage{bbding}
\usepackage[table,xcdraw]{xcolor}
\usepackage{float}
\usepackage{blindtext}

\title{SemanticBEVFusion: Rethink LiDAR-Camera Fusion in Unified Bird’s-Eye View Representation for 3D Object Detection}

\author{
 Qi Jiang \thanks{indicates equal contributions.}\\
  Bosch Corporate Research\\
  Shanghai Jiao Tong University\\
   \And
 Hao Sun\textsuperscript{*}\\
  Bosch Corporate Research\\
  \And
 Xi Zhang \\
  Shanghai Jiao Tong University\\
}

\begin{document}
\maketitle
\begin{abstract}
LiDAR and camera are two essential sensors for 3D object detection in autonomous driving. LiDAR provides accurate and reliable 3D geometry information while the camera provides rich texture with color. Despite the increasing popularity of fusing these two complementary sensors, the challenge remains in how to effectively fuse 3D LiDAR point cloud with 2D camera images. Recent methods focus on point-level fusion which paints the LiDAR point cloud with camera features in the perspective view or bird's-eye view (BEV)-level fusion which unifies multi-modality features in the BEV representation. In this paper, we rethink these previous fusion strategies and analyze their information loss and influences on geometric and semantic features. We present SemanticBEVFusion to deeply fuse camera features with LiDAR features in a unified BEV representation while maintaining per-modality strengths for 3D object detection. Our method achieves state-of-the-art performance on the large-scale nuScenes dataset, especially for challenging distant objects. The code will be made publicly available.
\end{abstract}


\section{Introduction}
\label{sec:intro}

3D object detection is one of the fundamental tasks in autonomous driving, which locates a set of objects in 3D space and infers their dimensions with categories. Nowadays, autonomous driving cars are equipped with diverse sensors including cameras, LiDAR, and so on, where different sensors are complementary to each other. LiDAR provides accurate and reliable 3D geometry information but lacks rich texture, meanwhile camera provides rich texture with color but is sensitive to illumination conditions. Meanwhile, different sensors have different modalities: the camera image is in perspective view and the LiDAR point cloud is in 3D view. Efficient fusion across different modalities has been a hot research field and is essential for accurate and reliable perception.

\begin{figure}[t!]
\centering
\includegraphics[width=\textwidth]{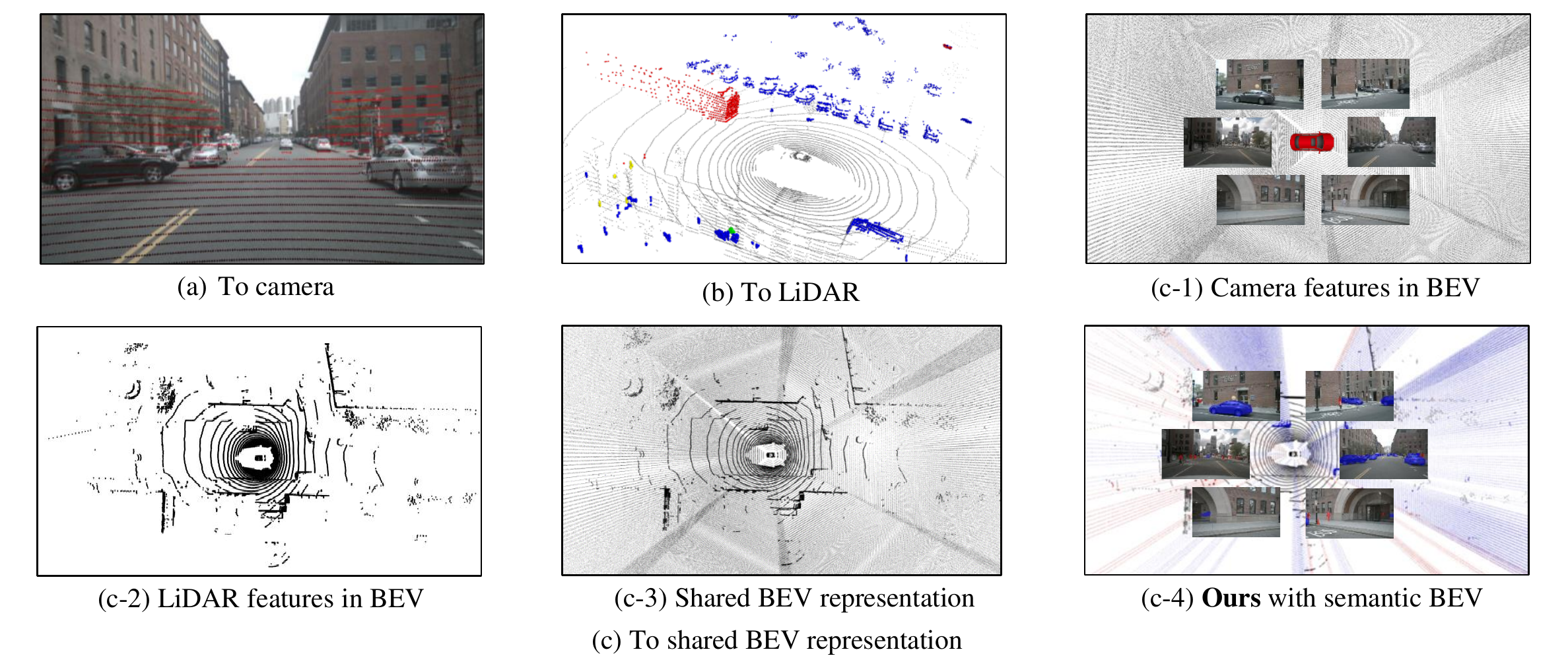}
\caption{\textbf{Fusion strategy comparison}. (a) To camera \cite{chen2017multi, gupta2014learning, deng2017amodal, alexandre20163d} projects LiDAR points to image and regresses 3D detection based on RGB-D data; (b) To LiDAR \cite{vora2020pointpainting, yin2021multimodal, wang2021pointaugmenting, fusionpainting2021, deepfusion2022} decorates LiDAR points with camera features for 3D detection; (c) BEVFusion\cite{liu2022bevfusion, liang2022bevfusion} fuses camera (c-1) and LiDAR (c-2) in one unified BEV representation (c-3). (c-4) Our SemanticBEVFusion maintains modality-specific strengths in one single BEV representation for boosted performance.}
\label{Fusion comparison}
\end{figure}

Benefiting from the success in 2D perception, early methods \cite{chen2017multi, gupta2014learning, deng2017amodal, alexandre20163d} project the LiDAR point cloud onto the image plane and perform 2D to 3D inference based on the RGB image with depth, as shown in Fig.\ref{Fusion comparison} (a). Later, point-level fusion methods \cite{vora2020pointpainting, yin2021multimodal, wang2021pointaugmenting, fusionpainting2021, deepfusion2022} paint LiDAR point cloud with 2D camera features, and use LiDAR backbone for 3D detection, as shown in Fig.\ref{Fusion comparison} (b).  Recent BEVFusion methods \cite{liu2022bevfusion, liang2022bevfusion} unify multi-modal features in a common BEV space by projecting multi-view image features to the 3D representation and fusing with LiDAR features in the BEV representation, as shown in Fig.\ref{Fusion comparison} (c1-3). BEVFusion methods\cite{liu2022bevfusion, liang2022bevfusion} outperform the point-level fusion methods and achieve state-of-the-art performance on public dataset benchmarks.

BEVFusion methods serve as an elegant and strong baseline in the research trend. However, transforming different modality representations into one single unified representation can lead to loss of modality-specific representational strengths \cite{yang2022deepinteraction}. Moreover, the contributions of each modality feature to the pipeline have not been well-studied. In this paper, we rethink the BEVFusion strategies and analyze the modality-specific influence on 3D geometric and semantic inference. We present an approach to maintain modality-specific strengths in one single fused BEV representation for 3D object detection, as shown in Fig.\ref{Fusion comparison} (c4). It boosts the LiDAR-only backbone by 5.4$\%$ mAP. Without bells and whistles, our framework achieves state-of-the-art performance on the popular nuScenes 3D object detection benchmark, outperforming state-of-the-art BEVFusion \cite{liu2022bevfusion} by 0.7$\%$ mAP and 0.1$\%$ NDS, especially 1.53$\%$ and 1.04$\%$  mAP and NDS in far range; BEVFusion \cite{liang2022bevfusion} by 1.7$\%$ mAP and 1.2$\%$ NDS.

Our contributions are summarized as follows: (i) We rethink the current BEV fusion strategies \cite{liu2022bevfusion,liang2022bevfusion}and analyze the semantic and geometric representational characteristics of LiDAR features and camera features on 3D detection. We claim and prove the necessity of semantic fusion in view transformation. (ii) We present a simple and efficient approach to maintain modality-specific strengths in one single fused BEV representation for 3D object detection. Our work can serve as a general fusion strategy for other approaches. (iii) Extensive experiments show that our approach outperforms the state-of-the-art methods on the nuScenes dataset, which validates its effectiveness. 

\section{Related Work}
\label{sec:Related Work}

{\bf LiDAR-Based 3D Object Detection} targets to infer three-dimensional rotated bounding boxes with corresponding categories in LiDAR point cloud \cite{yang2018pixor, bai2022transfusion, zhou2020end, sun20183d, zhou2018voxelnet, lang2019pointpillars, yang2018pixor, yan2018second}. Due to the unordered, irregular nature of
point cloud, many 3D detectors first encode them to a
regular grid such as 3D voxel \cite{zhou2018voxelnet} and pillar \cite{lang2019pointpillars} representations, then point cloud features are computed by standard 2D or 3D convolutions \cite{qi2017pointnet, graham20183d} for detection predictions. VoxelNet \cite{zhou2018voxelnet} voxelizes the point cloud in 3D space and applies PointNet \cite{qi2017pointnet} inside each voxel for feature representation generation, then 3D sparse convolutions \cite{graham20183d} and bird-eye view 2D convolutions are utilized with 2D detection head for regressing detections. PointPillars method \cite{lang2019pointpillars} replaces the voxel computation by an elongated pillar representation and detections are regressed in BEV, similar to PIXOR \cite{yang2018pixor}.
SECOND \cite{yan2018second} improves the efficiency of sparse 3D convolutions. Following 2D detection \cite{ren2015faster, redmon2016you}, the mainstream 3D detection methods use anchor-based 3D detection head, while CenterPoint \cite{yin2021center} adopts a center-based representation for anchor-free 3D object detection.

{\bf Camera-Based 3D Object Detection} predicts 3D rotated bounding boxes with corresponding semantic information from camera images. Mono3D \cite{chen2016monocular} makes the ground-plane assumption for 3D detection. CenterNet \cite{duan2019centernet} performs 2D detection in the image plane and predicts 3D bounding boxes based on center features. FCOS3D \cite{wang2021fcos3d} extends 2D detector \cite{tian2019fcos} with additional 3D regression branches. DETR3D \cite{wang2022detr3d}, PETR \cite{liu2022petr} learns object queries in the 3D space for detection instead of in the perspective view. Pseudo-LiDAR methods \cite{huang2021bevdet} convert the camera features from perspective view to the bird’s-eye view using a view transformer such as LSS \cite{philion2020lift} and OFT \cite{roddick2018orthographic}. Pure vision-based methods are outperformed by LiDAR and LiDAR-Camera fusion-based methods.

{\bf LiDAR-Camera Object Detection} has gained increasing attention in the 3D detection
community. Early works such as MV3D \cite{chen2017multi} and F-PointNet \cite{qi2018frustum} perform proposal-level fusion, where features inside RoI are coarsely fused across two modalities. Point-level fusion methods such as PointPainting \cite{vora2020pointpainting}, MVP \cite{yin2021multimodal}, PointAugmenting \cite{wang2021pointaugmenting} paint camera features onto LiDAR points and perform LiDAR-based detection
on the augmented point cloud afterward. TransFusion \cite{bai2022transfusion} learns object queries in the 3D space and fuses image features onto object-level proposals.  BEVFusion methods \cite{liu2022bevfusion, liang2022bevfusion} utilize LSS \cite{philion2020lift} to transform image features from perspective view to BEV and fuses vision BEV features with LiDAR features in the unified representation. While BEV fusion methods\cite{liu2022bevfusion, liang2022bevfusion} have achieved outstanding performance and become a hot fusion trend, we rethink the fusion logic and analyze how each modality works/helps in the detection task. Based on that, we propose our approach: SemanticBEVFusion which further carries on each modality representational strength into one elegant BEV fusion and improves current BEVFusion performance.

\section{Methodology}
\label{sec:Method}

\begin{figure*}[t!]
\centering
\includegraphics[width=\textwidth]{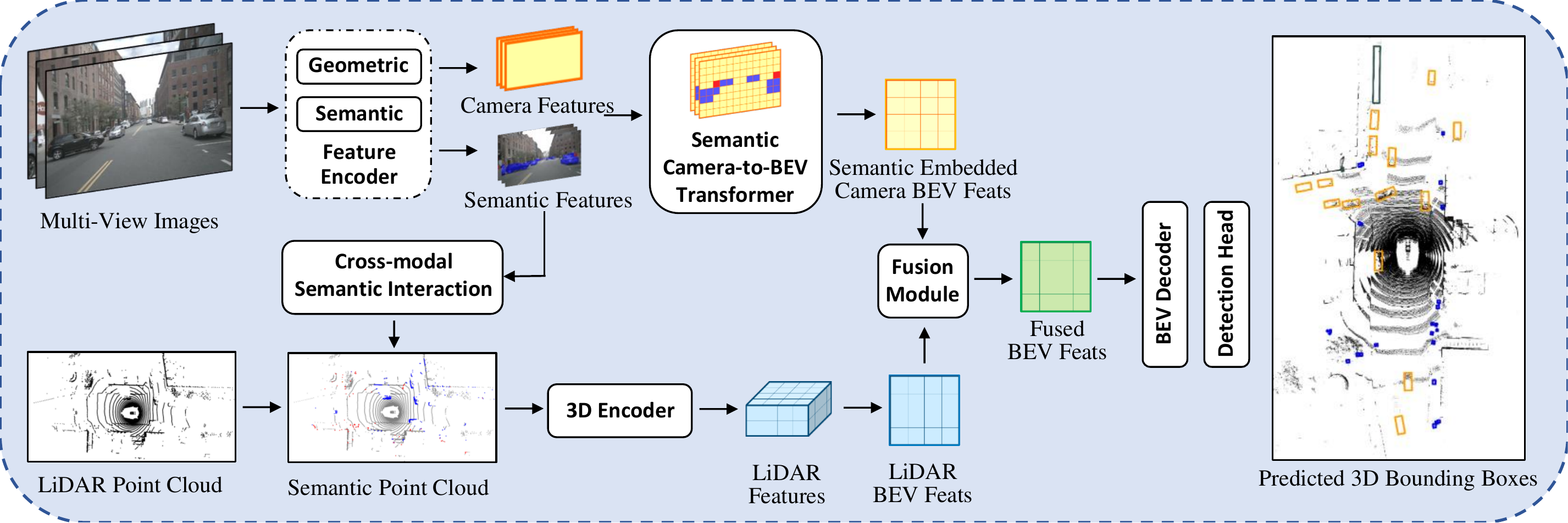}
\caption{\textbf{An overview of SemanticBEVFusion}. Our approach consists of two concurrent sensor streams and a shared BEV decoding: a). Semantic features are inferred based on the rich textures of images and fused with camera geometric features. Semantic decorated camera features of foreground objects are transformed into the BEV representation. b). The 2D semantic features are transferred to LiDAR point cloud and an off-the-shelf 3D backbone is used to extract LiDAR BEV features. c). Semantic embedded camera BEV features and LiDAR BEV features are fused for detection heads.}
\label{main method pipeline}
\end{figure*}

In this section, we present the details of our LiDAR-camera fusion 3D detection approach. As shown in Fig.\ref{main method pipeline}, our model mainly consists of two concurrent LiDAR-camera streams and a shared BEV decoder based on transformer. We rethink the prior BEVFusion works \cite{liu2022bevfusion, liang2022bevfusion} and discuss the contributions of semantic and geometric features to detection performance. We propose a novel semantic-geometric interaction mechanism for both streams to maintain the modality-specific strengths for boosted performance. 

\subsection{Camera Stream}
\label{sec:camera stream}
The camera stream pipeline is shown in the upper part of Fig.\ref{main method pipeline}. The camera stream can be decomposed into two fundamental modules:

\textbf{Image Feature Encoding} encodes each raw image, $I\in R^{3\times H_I\times W_I}$, where $H_I$ and $W_I$ are the height and width of the input images, into explicit semantic features and geometry-rich deep features on perspectives view:

\begin{itemize}
  \item \textbf{Semantic Features} 
  The camera image is rich in semantic attributes. We explicitly encode its semantic representational features by instance segmentation. Each image is input to CenterNet2\cite{zhou2021probablistic} with DLA34\cite{yu2018deep} and cascade ROI heads to obtain instance segmentation masks $M$ over $N$ object categories.

  \item \textbf{Geometric Features} 3D geometric attributes are latent and implicit in the image.  We use Swin-Tiny\cite{liu2021Swin} to encode the geometric features of each image as
  $F^C\in R^{C_{cam}\times H_F\times W_F}$, where $C_{cam}$ is the feature channels, $H_F$ and $W_F$ are the height and width.
  
  \item \textbf{Semantic Embedded Features} Previous works \cite{liu2022bevfusion, liang2022bevfusion} transform geometric features $F^C$ in perspective view to 3D coordinate with inferred discrete depth estimation as shown in Fig.\ref{semantic painting comparison} (a). However, we claim that besides inferring deep features with latent geometric attributes from the image, fusing geometric features with explicit semantic features can further boost performance, especially for small and distant objects. We project the instance segmentation masks $M$ to the dimension of $F^C$ as $M^S$ and fuse with $F^C$ through a multi-layer perception (MLP) and element-wise addition for semantic embedded camera features $F^{CS}\in R^{C_{cam}\times H_F\times W_F}$.

\end{itemize}

\begin{figure*}[t!]
\centering
\includegraphics[width=\textwidth]{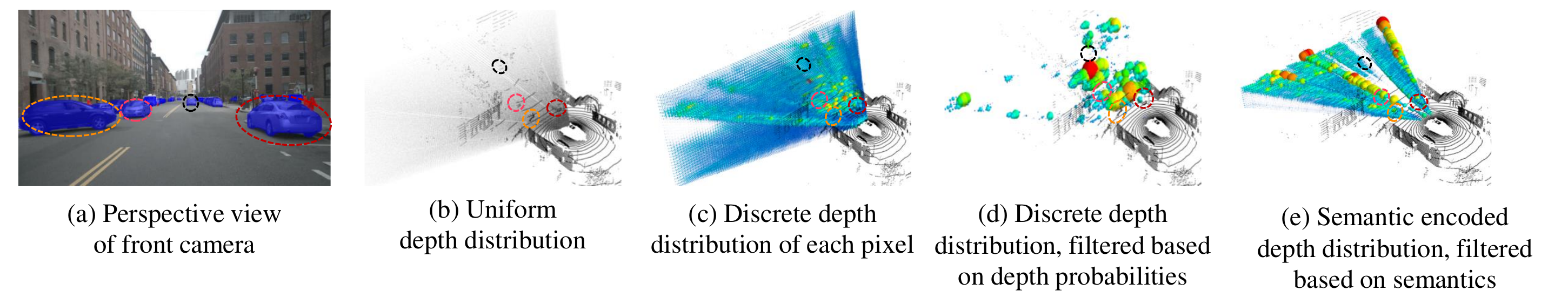}
\caption{\textbf{Camera BEV features comparison.} (a). The front camera image is shown where different traffic participants are inferred in blue for vehicles and red for other categories. Four cars at varying distances are highlighted by colored circles in the perspective view and the following 3D views. (b). Uniform depth distribution without any depth probability prediction. (c). Discrete depth distribution prediction \cite{philion2020lift, liu2022bevfusion, huang2021bevdet}. Pseudo points with high probabilities are marked in red and large, and vice versa. (d). Discrete depth distribution, pseudo points with depth probabilities larger than a certain threshold are shown.  (e). Our semantic encoded depth distribution, pseudo points with semantic probabilities larger than a certain threshold are shown. }
\label{semantic painting visualize}
\end{figure*}

\textbf{Semantic View Transformation} transforms semantic embedded camera features in perspective view to the BEV representation, as shown in Fig.\ref{semantic painting comparison} (b) and (c). Compared with \cite{philion2020lift, liu2022bevfusion, liang2022bevfusion}, we further decompose it into three key steps to have a deeper understanding:

\begin{itemize}

  \item \textbf{Mapping} scatters the 2D features from perspective view to BEV view based on uniform depth distribution.
  The per-pixel camera feature is associated with a set of 3D pseudo points which are evenly distributed along the depth direction based on extrinsic and intrinsic parameters.
 As shown in Fig.\ref{semantic painting visualize} (b), the mapping process guarantees the dense scattering of the camera features along the ray direction, covering all the possible locations of objects which can be accessible for later fusion to the LiDAR features at approximate BEV locations.
\begin{figure}[t!]
\centering
\includegraphics[width=\textwidth]{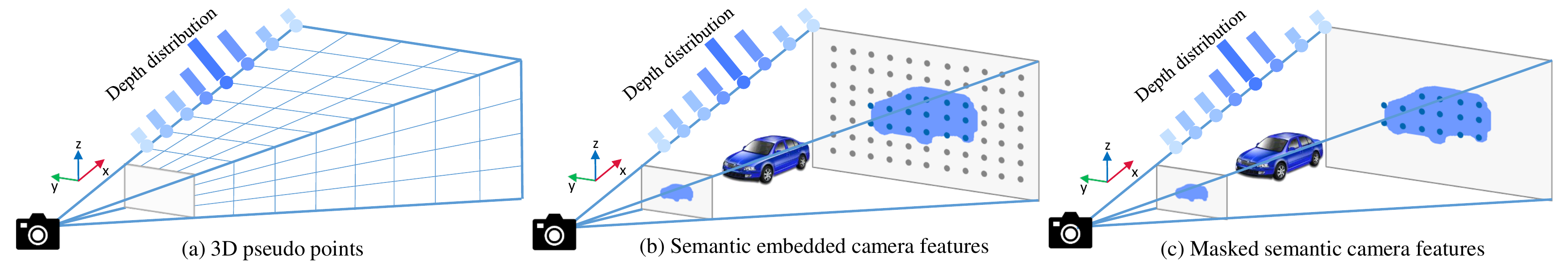}
\caption{\textbf{An illustration of different camera-BEV view transformation mechanisms. }
(a). Pseudo points with discrete depth distribution probabilities along the ray are commonly used for camera-BEV transformation\cite{philion2020lift, liu2022bevfusion, liang2022bevfusion, xie2022m, huang2021bevdet, huang2022bevdet4d, li2022bevdepth}. (b). Semantic features are inferred and encoded to camera geometric features for enhanced semantic supervision. (c). Foreground camera features with task-related semantics are selectively transformed to BEV, which makes pseudo point processing more efficient. 
}
\label{semantic painting comparison}
\end{figure}
  \item \textbf{Depth Attention} adjusts the weights of per-pixel camera features  along the ray direction.
  The per-pixel camera feature is used to predict a context vector parameterized by discrete depth inference to match a notion of attention\cite{philion2020lift}, strengthening or weakening the features along the ray direction. This depth attention process exploits latent depth distribution to attend to the pseudo points with highly possible locations of the objects as illustrated in Fig.\ref{semantic painting visualize} (c). Although it has been proven effective in LSS\cite{philion2020lift},  CaDNN\cite{CaDDN} with supervised depth prediction, the importance of depth accuracy for the camera feature is diminished in the LiDAR-camera fusion method. We surprisingly find that even with only mapping (uniform depth distribution), our SemanticBEVFusion boosts the LiDAR only backbone by 3.4$\%$ mAP and achieves relatively similar performances compared with adding predicted depth distribution \cite{liu2022bevfusion, liang2022bevfusion} (Table. \ref{view transformation}). We rethink the view transformation mechanisms in the prior methods\cite{liu2022bevfusion, liang2022bevfusion} and bring up the question: which parts of camera features lift up the camera-LiDAR fusion more: the semantics or the 3D pseudo geometric prediction?

 \item \textbf{Semantic Masking}  Camera pesudo points are dense and expensive to compute \cite{philion2020lift, liu2022bevfusion, liang2022bevfusion}.
    We propose to filter the background features based on semantic masks to preserve camera features of foreground objects $F_{BEV}^{CS}\in R^{C_{con} \times H_B \times W_B}$, where $H_B$ and $W_B$ are the resolutions of BEV features, as illustrated in Fig.\ref{semantic painting comparison} (c).
      The masking process significantly reduces the number of pseudo points by 84.36$\%$ and preserves valuable camera features for later fusion as shown in Fig.\ref{semantic painting visualize} (c) and (e). 
      Filtering based on the depth distribution also reduces the number of pseudo points as in Fig.\ref{semantic painting visualize} (d) but potentially undermines the spatial correspondence between camera BEV features and LiDAR BEV features, causing a performance drop of 0.62$\%$ in mAP. Please refer to Table.\ref{component analysis}  row \emph{e} and \emph{f} for more details. Compared with depth attention probabilities, camera semantic features dominate the object proposal generation.   

\end{itemize}

\subsection{LiDAR Stream}
\label{sec:lidar stream}
The LiDAR stream pipeline is shown in the bottom part of Fig.\ref{main method pipeline}.  Similar to \cite{vora2020pointpainting, yin2021multimodal}, we fuse the semantic features from camera with LiDAR points $P=\{p_i=(x,y,z,t,i) \in R^5\}$ to generate semantic LiDAR points $P^S=\{p^S_i=(x,y,z,t,i,c,s) \in R^{5+N+1}\}$, over $N$ categories with its confidence score $s$.  The fusion is conducted between LiDAR points and 2D image segmentation instances based on intrinsic and extrinsic parameters, sufficiently transferring semantic information to the shared BEV representation. These LiDAR points decorated with segmentation information can be easily modulated into any off-the-shelf 3D detection models. Following \cite{zhou2018voxelnet}, the semantic points are voxelized into voxel features and we use sparse 3D convolution\cite{yan2018second} to encode as semantic LiDAR BEV features $F_{BEV}^{LS}$. This fusion mechanism benefits object with sparse point cloud features by dense image features for improved object proposals with better classification and dimension estimation.

\subsection{Fusion Module}
\label{sec:fusion module}
Given the semantic embedded camera features and LiDAR features in the shared BEV representation, following \cite{liu2022bevfusion}, we adapt a convolutional BEV encoder for fusion as described below:

\begin{equation}
F_{BEV}^{fused} = f_{conv}(f_{cat}(F_{BEV}^{CS}, F_{BEV}^{LS}))
\end{equation}

We also test an element-wise additive encoder for fusion, formulated as:
\begin{equation}
F_{BEV}^{fused} = f_{conv}(F_{BEV}^{CS}) \oplus  f_{conv}(F_{BEV}^{LS})
\end{equation}

We observe similar performances of two fusion mechanisms (Table.\ref{component analysis}), though methods \cite{liu2022bevfusion} claim that a well-designed BEV fusion module is necessary to compensate for the spatial misalignment of LiDAR BEV features and camera BEV features, which is caused by the inaccurate depth estimation during view transformation. We think that compared with the 3D pseudo geometric features predicted by the depth attention, camera semantic features dominate the overall BEVFusion performance, proving the effectiveness of semantic fusion.

\subsection{Detection Head}
\label{sec:detection head}
Popular 3D perception heads can be easily modulated into our framework. We follow the design of DETR3D\cite{wang2022detr3d} and TransFusion\cite{bai2022transfusion} to use transformers for global feature aggregation. The query features are input into the anchor-free detection head\cite{yin2021center} to predict the object dimension, orientation, semantic category, and other attributes.


\section{Experiments}
\label{sec:Experiments}
\subsection{Experimental Settings}
\label{sec:experimental settings}

In this section, we present the performance of our method with experimental settings on the challenging nuScenes benchmark. Ablation studies are conducted for a better understanding of the designed components.

\noindent{\textbf{Dataset and Metrics.}}
The nuScenes dataset \cite{nuscenes2019} is one of the largest multimodal autonomous-driving datasets with 360$^{\circ}$ sensor coverage based on the sensor suite including cameras, LiDARs, and so on.  The dataset contains 1000 driving sequences, with each 20s long and annotated with 3D bounding boxes for 10 classes: cars, trucks, buses,
trailers, construction vehicles, pedestrians, motorcycles, bicycles, traffic cones, and barriers. We follow the official dataset split to use 700, 150, and 150 sequences for training, validation, and testing. For 3D object detection, we follow the official evaluation metrics including mean Average Precision (mAP) and nuScenes detection score (NDS). Please refer to\cite{nuscenes2019} for more details regarding metrics definition.  

\noindent{\textbf{Implementation Details.}}
Our method is implemented based on the open-sourced MMdetection3d \cite{mmdet3d2020} for 3D detection and CenterNet2\cite{zhou2021probablistic} for 2D instance segmentation in Pytorch. The CenterNet2 is pre-trained on COCO and fine-tuned on NuImages and its weights are frozen during training. We set the image size to 512$\times$928 for CenterNet2\cite{zhou2021probablistic} and 256$\times$704 for Swin-Tiny\cite{liu2021Swin}.We follow a two-stage training manner \cite{bai2022transfusion, liang2022bevfusion}: i). We train the LiDAR stream with semantic points for 20 epochs, adopting the common data augmentation and training schedules of prior LiDAR methods\cite{bai2022transfusion, yin2021center}. Fade strategy\cite{wang2021pointaugmenting} is utilized in the last 5 epochs to adapt the model to the genuine distribution. ii). We then train the whole SemanticBEVFusion pipeline with a pre-initialized camera stream, inheriting weights from the LiDAR stream for additional 6 epochs.  We use adamW optimizer with a one-cycle learning rate policy, max learning of 1e-3, weight decay of 0.1, and momentum between 0.85 and 0.95. We conduct our experiments on 4 Nvidia GTX 3090 GPUs with a batch size of 4 for 1st stage training and 2 for 2nd stage training.

\begin{figure*}[t!]
\centering
    \includegraphics[width=1.0\textwidth]{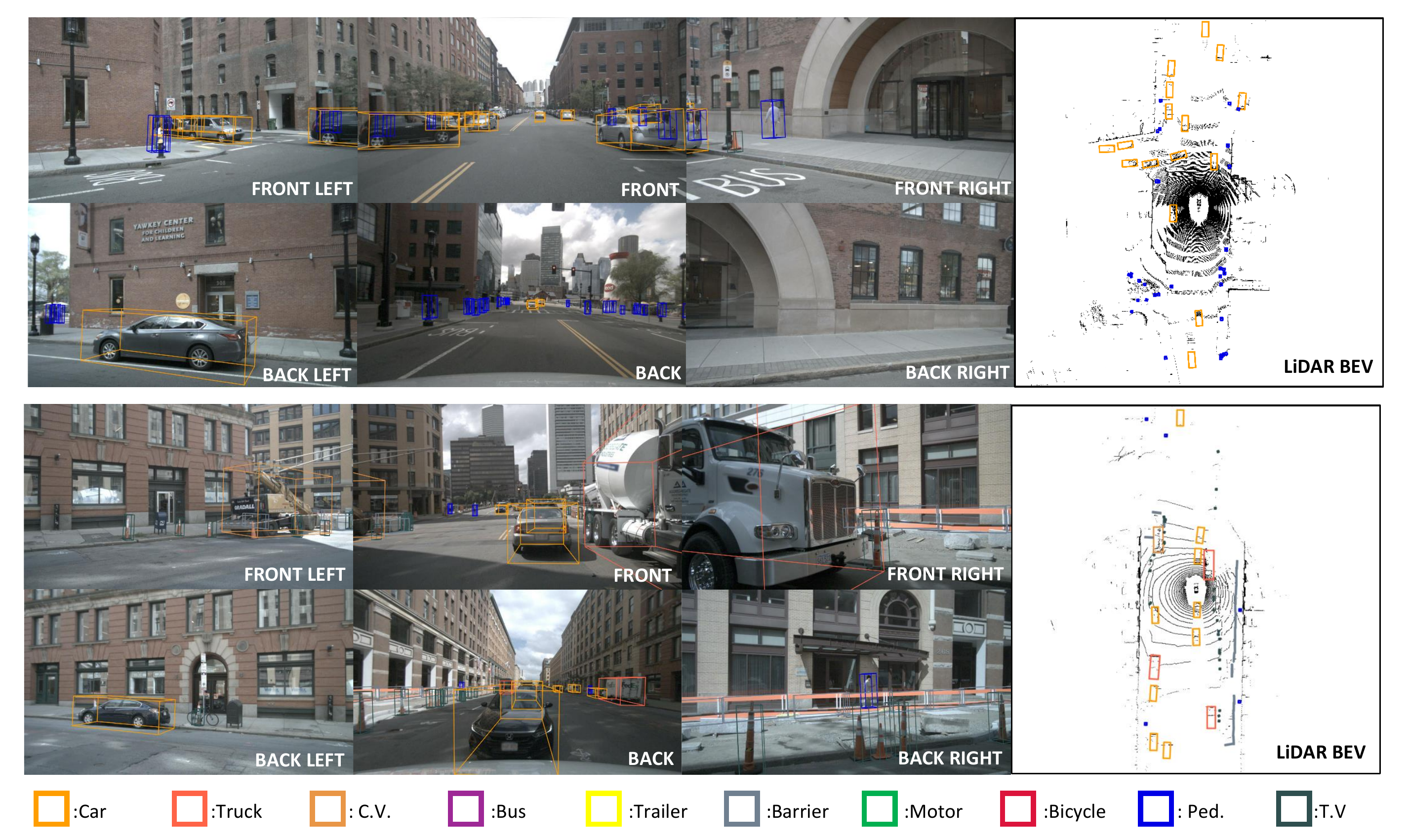}
    \caption{Qualitative results on nuScenes val set. Predicted results are marked with different colors for different categories. Our model accurately detects crowded cars with different degrees of occlusion problems (top). A more diverse scene with various object categories, especially small objects like traffic cones, bicycles, and pedestrians is also shown (bottom).}
    \label{Qualitative results}
\end{figure*}

\subsection{Comparison with State-of-the-arts}
\label{sec:test sota}
\begin{table*}[t!]
    \caption{\textbf{3D object detection on nuScenes val set (top) and test set (bottom) across different categories.} Our method achieves state-of-the-art  performance, especially on small objects like pedestrians and traffic cones. (C.V., Ped., and T.C. are abbreviated for construction vehicle, pedestrian, and traffic cone.  \dag: test-time augmentation (TTA). \ddag: Our re-implementation.)}
\centering
\resizebox{\textwidth}{!}{
    \begin{tabular}{l|c|cc|cccccccccc}
    \toprule
    Method & Modality & mAP   & NDS   & Car   & Truck & Bus   & Trailer & C.V.  & Ped.  & Motor & Bike  & T.C.  & Barrier \\
        \midrule
    TransFusion-L\cite{bai2022transfusion}$^{\ddag}$  & L     & 64.2  & 69.2  & 87.0  & 59.7  & 72.9  & 42.1  & 24.9  & 86.2  & 70.9  & 55.3  & 73.4  & 69.7 \\
    BEVFusion\cite{liu2022bevfusion} & L+C   & 68.5  & 71.4  & 89.3  & 64.9  & \textcolor[rgb]{ 0,  0,  1}{\textbf{75.7}} & 43.6  & 29.8  & 88.2  & \textcolor[rgb]{ 0,  0,  1}{\textbf{78.6}} & 64.1  & 79.5  & 71.0 \\
    \textbf{SemanticBEV (Ours)}  & L+C   & \textcolor[rgb]{ 0,  0,  1}{\textbf{69.5}} & \textcolor[rgb]{ 0,  0,  1}{\textbf{72.0}} & \textcolor[rgb]{ 0,  0,  1}{\textbf{89.5}} & \textcolor[rgb]{ 0,  0,  1}{\textbf{65.9}} & 75.3  & \textcolor[rgb]{ 0,  0,  1}{\textbf{44.3}} & \textcolor[rgb]{ 0,  0,  1}{\textbf{30.5}} & \textcolor[rgb]{ 0,  0,  1}{\textbf{90.0}} & 78.0  & \textcolor[rgb]{ 0,  0,  1}{\textbf{66.0}}  & \textcolor[rgb]{ 0,  0,  1}{\textbf{81.7}} & \textcolor[rgb]{ 0,  0,  1}{\textbf{73.5}} \\
    \midrule
    \midrule
    CenterPoint\cite{yin2021center}$^{\dag}$ & L     & 60.3  & 67.3  & 85.2  & 53.5  & 63.6  & 56.0  & 20.0  & 84.6  & 59.5  & 30.7  & 78.4  & 71.1 \\
    TransFusion-L\cite{bai2022transfusion} & L     & 65.5  & 70.2  & 86.2  & 56.7  & 66.3  & 58.8  & 28.2  & 86.1  & 68.3  & 44.2  & 82.0  & 78.2 \\
    \midrule
    PointPainting\cite{vora2020pointpainting} & L+C   & 46.4  & 58.1  & 77.9  & 35.8  & 36.2  & 37.3  & 15.8  & 73.3  & 41.5  & 24.1  & 62.4  & 60.2 \\
    MVP\cite{yin2021multimodal}   & L+C   & 66.4  & 70.5  & 86.8  & 58.5  & 67.4  & 57.3  & 26.1  & 89.1  & 70.0  & 49.3  & 85.0  & 74.8 \\
    PointAugmenting\cite{wang2021pointaugmenting}$^{\dag}$ & L+C   & 66.8  & 71.1  & 87.5  & 57.3  & 65.2  & 60.7  & 28.0  & 87.9  & 74.3  & 50.9  & 83.6  & 72.6 \\
    FusionPainting\cite{fusionpainting2021} & L+C   & 68.1  & 71.6  & 87.1  & 60.8  & 68.5  & 61.7  & 30.0  & 88.3  & 74.7 & 53.5 & 85.0  & 71.8 \\
    TransFusion\cite{bai2022transfusion} & L+C   & 68.9  & 71.7  & 87.1  & 60.0  & 68.3  & 60.8  & 33.1  & 88.4  & 73.6  & 52.9  & 86.7  & 78.1 \\
    BEVFusion\cite{liang2022bevfusion} & L+C   & 69.2  & 71.8  & 88.1  & 60.9  & 69.3  & 62.1  & 34.4  & 89.2  & 72.2  & 52.2  & 85.2  & 78.2 \\
    BEVFusion\cite{liu2022bevfusion} & L+C   & 70.2  & 72.9  & 88.6  & 60.1  & 69.8  & \textcolor[rgb]{ 0,  0,  1}{\textbf{63.8}} & \textcolor[rgb]{ 0,  0,  1}{\textbf{39.3}} & 89.2  & 74.1  & 51.0  & 86.5  & \textcolor[rgb]{ 0,  0,  1}{\textbf{80.0}} \\
    \textbf{SemanticBEV (Ours)} & L+C & \textcolor[rgb]{ 0,  0,  1}{\textbf{70.9}} & \textcolor[rgb]{ 0,  0,  1}{\textbf{73.0}} & \textcolor[rgb]{ 0,  0,  1}{\textbf{89.1}} & \textcolor[rgb]{ 0,  0,  1}{\textbf{61.5}} & \textcolor[rgb]{ 0,  0,  1}{\textbf{72.3}} & 63.7  & 39.2  & \textcolor[rgb]{ 0,  0,  1}{\textbf{90.9}} & \textcolor[rgb]{ 0,  0,  1}{\textbf{74.7}} & \textcolor[rgb]{ 0,  0,  1}{\textbf{53.5}} & \textcolor[rgb]{ 0,  0,  1}{\textbf{87.4}} & 77.2 \\
    \bottomrule
    \end{tabular}
    }
\label{across different categories}
\end{table*}

\begin{table*}[htbp]
\caption{\textbf{3D object detection with true positive metrics on nuScenes (val) over different ranges.} Higher AP, NDS and lower ATE, ASE, AOE, AVE, AAE indicate better performance, marked in blue. Ranges are defined by the centric distances between ego car and objects. Our method achieves state-of-the-art performance, especially in far range. ($^*$: Our reproduced model of BEVFusion-L\cite{liu2022bevfusion}, also know as TransFusion-L\cite{bai2022transfusion}))}
  \centering
    \resizebox{\textwidth}{!}{
    \begin{tabular}{llccccccc}
    \toprule
    Range & Method & mAP↑ & NDS↑ & mATE↓  & mASE↓  & mAOE↓  & mAVE↓  & mAAE↓ \\
    \midrule
    
    \multirow{4}[2]{*}{Whole} 
    & LiDAR-only $^*$\cite{liu2022bevfusion, bai2022transfusion}
    & 64.24
    & 69.19
    & 29.03
    & 25.27
    & \cellcolor[rgb]{ .906,  .902,  .902}\textbf{29.51}
    & 26.33
    & 19.15\\
    
      & BEVFusion \cite{liu2022bevfusion}
      & 68.45 \scriptsize{\textcolor[rgb]{ 0,  0,  1}{(+4.21)}}
      & 71.42 \scriptsize{\textcolor[rgb]{ 0,  0,  1}{(+2.23)}}
      & 28.77 \scriptsize{\textcolor[rgb]{ 0,  0,  1}{(-0.26)}}
      & 25.37 \scriptsize{\textcolor[rgb]{ 1,  0,  0}{(+0.10)}}
      & 29.83 \scriptsize{\textcolor[rgb]{ 1,  0,  0}{(+0.32)}}
      & 25.25 \scriptsize{\textcolor[rgb]{ 0,  0,  1}{(-1.08)}}
      & \cellcolor[rgb]{ .906,  .902,  .902}\textbf{18.78} \scriptsize{\cellcolor[rgb]{ .906,  .902,  .902}\textcolor[rgb]{ 0,  0,  1}{\textbf{(-0.37)}}} \\

      & \textbf{SemanticBEV (Ours)} 
      & \cellcolor[rgb]{ .906,  .902,  .902}\textbf{69.47} \scriptsize{\cellcolor[rgb]{ .906,  .902,  .902}\textcolor[rgb]{ 0,  0,  1}{\textbf{(+5.23)}} }
      & \cellcolor[rgb]{ .906,  .902,  .902}\textbf{71.96} \scriptsize{\cellcolor[rgb]{ .906,  .902,  .902}\textcolor[rgb]{ 0,  0,  1}{\textbf{(+2.77)}} }
      &  \cellcolor[rgb]{ .906,  .902,  .902}\textbf{27.94} \scriptsize{\textcolor[rgb]{ 0,  0,  1}{\textbf{(-1.09)}} }
      & \cellcolor[rgb]{ .906,  .902,  .902}\textbf{24.90} \scriptsize{ \cellcolor[rgb]{ .906,  .902,  .902}\textcolor[rgb]{ 0,  0,  1}{\textbf{(-0.37)}} }
      & 31.65 \scriptsize{ \textcolor[rgb]{ 1,  0,  0}{(+2.14)} }
      & \cellcolor[rgb]{ .906,  .902,  .902}\textbf{24.31} \scriptsize{ \cellcolor[rgb]{ .906,  .902,  .902}\textcolor[rgb]{ 0,  0,  1}{\textbf{(-2.02)}} }
      & 18.97 \scriptsize{\textcolor[rgb]{ 0,  0,  1}{(-0.18)}} \\
          
        \midrule
        \multirow{4}[1]{*}{Near} 
        & LiDAR-only $^*$\cite{liu2022bevfusion, bai2022transfusion}
        & 77.44
        & 76.30
        & 22.90
        & 24.59
        & 24.15
        & 23.64
        & 28.94
        \\
          & BEVFusion \cite{liu2022bevfusion}
          & 78.41 \scriptsize{\textcolor[rgb]{ 0,  0,  1}{(+0.97)} }
          & 77.16 \scriptsize{\textcolor[rgb]{ 0,  0,  1}{(+0.86)} }
          & 22.75 \scriptsize{\textcolor[rgb]{ 0,  0,  1}{(-0.15)} }
          & 24.50 \scriptsize{\textcolor[rgb]{ 0,  0,  1}{(-0.09)} }
          & 24.41 \scriptsize{\textcolor[rgb]{ 1,  0,  0}{(+0.26)} }
          & 22.97 \scriptsize{\textcolor[rgb]{ 0,  0,  1}{(-0.67)} }
          & \cellcolor[rgb]{ .906,  .902,  .902}\textbf{25.82} \scriptsize{\cellcolor[rgb]{ .906,  .902,  .902}\textcolor[rgb]{ 0,  0,  1}{\textbf{(-3.12)}} }
          \\

          & \textbf{SemanticBEV (Ours)} 
          & \cellcolor[rgb]{ .906,  .902,  .902}\textbf{79.85} \scriptsize{ \cellcolor[rgb]{ .906,  .902,  .902}\textcolor[rgb]{ 0,  0,  1}{\textbf{(+2.41)}} }
          & \cellcolor[rgb]{ .906,  .902,  .902}\textbf{78.08} \scriptsize{ \cellcolor[rgb]{ .906,  .902,  .902}\textcolor[rgb]{ 0,  0,  1}{\textbf{(+1.78)}} }
          &  \cellcolor[rgb]{ .906,  .902,  .902}\textbf{21.78} \scriptsize{ \textcolor[rgb]{ 0,  0,  1}{\textbf{(-1.12)}}}
          & \cellcolor[rgb]{ .906,  .902,  .902}\textbf{23.99} \scriptsize{ \textcolor[rgb]{ 0,  0,  1}{\textbf{(-0.60)}} }
          & \cellcolor[rgb]{ .906,  .902,  .902}\textbf{22.26} \scriptsize{ \textcolor[rgb]{ 0,  0,  1}{\textbf{(-1.89)}} }
          & \cellcolor[rgb]{ .906,  .902,  .902}\textbf{20.14} \scriptsize{ \cellcolor[rgb]{ .906,  .902,  .902}\textcolor[rgb]{ 0,  0,  1}{\textbf{(-3.50)}} }
          & 30.26 \scriptsize{ \textcolor[rgb]{ 1,  0,  0}{(+1.32)}} \\
    \midrule
    \multirow{4}[1]{*}{Middle} 
    & LiDAR-only $^*$\cite{liu2022bevfusion, bai2022transfusion}
    & 59.85
    & 67.19
    & 30.72
    & \cellcolor[rgb]{ .906,  .902,  .902}\textbf{25.33}
    & \cellcolor[rgb]{ .906,  .902,  .902}\textbf{30.64}
    & 26.25
    & 14.47
    \\
          & BEVFusion \cite{liu2022bevfusion}
          & 65.38 \scriptsize{\textcolor[rgb]{ 0,  0,  1}{(+5.53)} }
          & 69.90 \scriptsize{\textcolor[rgb]{ 0,  0,  1}{(+2.71)} }
          & 30.40 \scriptsize{\textcolor[rgb]{ 0,  0,  1}{(-0.32)} }
          & 25.45 \scriptsize{\textcolor[rgb]{ 1,  0,  0}{(+0.12)} }
          & 31.47 \scriptsize{\textcolor[rgb]{ 1,  0,  0}{(+0.83)} }
          & 25.24  \scriptsize{ \textcolor[rgb]{ 0,  0,  1}{(-1.01)} }
          & 15.33 \scriptsize{\textcolor[rgb]{ 1,  0,  0}{(+0.86)} }
          \\

          & \textbf{SemanticBEV (Ours)} 
          & \cellcolor[rgb]{ .906,  .902,  .902}\textbf{66.07} \scriptsize{ \cellcolor[rgb]{ .906,  .902,  .902}\textcolor[rgb]{ 0,  0,  1}{\textbf{(+6.22)}} }
          & \cellcolor[rgb]{ .906,  .902,  .902}\textbf{70.20} \scriptsize{ \cellcolor[rgb]{ .906,  .902,  .902}\textcolor[rgb]{ 0,  0,  1}{\textbf{(+3.01)}} }
          &  \cellcolor[rgb]{ .906,  .902,  .902}\textbf{29.46} \scriptsize{ \textcolor[rgb]{ 0,  0,  1}{\textbf{(-1.26)}} }
          & 25.42 \scriptsize{ \textcolor[rgb]{ 1,  0,  0}{(+0.09)} }
          & 35.26 \scriptsize{ \textcolor[rgb]{ 1,  0,  0}{(+4.62)} }
          & \cellcolor[rgb]{ .906,  .902,  .902}\textbf{24.00} \scriptsize{ \cellcolor[rgb]{ .906,  .902,  .902}\textcolor[rgb]{ 0,  0,  1}{\textbf{(-2.25)}} }
          & \cellcolor[rgb]{ .906,  .902,  .902}\textbf{14.20} \scriptsize{ \cellcolor[rgb]{ .906,  .902,  .902}\textcolor[rgb]{ 0,  0,  1}{\textbf{(-0.27)}} }
          \\
          
    \midrule
    \multirow{4}[2]{*}{Far} 
    & LiDAR-only $^*$\cite{liu2022bevfusion, bai2022transfusion}
    & 29.71
    & 44.94
    & 55.76
    & 41.75
    & 48.75
    & 34.66
    & 18.21
    \\
          
          & BEVFusion \cite{liu2022bevfusion}
          & 35.48 \scriptsize{ \textcolor[rgb]{ 0,  0,  1}{(+5.77)} }
          & 48.09 \scriptsize{ \textcolor[rgb]{ 0,  0,  1}{(+3.15)} }
          & 55.14 \scriptsize{ \textcolor[rgb]{ 0,  0,  1}{(-0.62)} }
          & 41.78 \scriptsize{ \textcolor[rgb]{ 1,  0,  0}{(+0.03)} }
          & 49.75 \scriptsize{ \textcolor[rgb]{ 1,  0,  0}{(+1.00)} }
          & \cellcolor[rgb]{ .906,  .902,  .902}\textbf{32.28} \scriptsize{ \cellcolor[rgb]{ .906,  .902,  .902}\textcolor[rgb]{ 0,  0,  1}{\textbf{(-2.38)}} }
          & \cellcolor[rgb]{ .906,  .902,  .902}\textbf{17.51} \scriptsize{ \cellcolor[rgb]{ .906,  .902,  .902}\textcolor[rgb]{ 0,  0,  1}{\textbf{(-0.70)}} }
          \\
          
          & \textbf{SemanticBEV (Ours)} 
          & \cellcolor[rgb]{ .906,  .902,  .902}\textbf{37.01} \scriptsize{ \cellcolor[rgb]{ .906,  .902,  .902}\textcolor[rgb]{ 0,  0,  1}{\textbf{(+7.30)}} }
          & \cellcolor[rgb]{ .906,  .902,  .902}\textbf{49.13} \scriptsize{ \cellcolor[rgb]{ .906,  .902,  .902}\textcolor[rgb]{ 0,  0,  1}{\textbf{(+4.19)}} }
          & \cellcolor[rgb]{ .906,  .902,  .902}\textbf{53.69} \scriptsize{ \cellcolor[rgb]{ .906,  .902,  .902}\textcolor[rgb]{ 0,  0,  1}{\textbf{(-2.07)}} }
          & \cellcolor[rgb]{ .906,  .902,  .902}\textbf{40.79} \scriptsize{ \cellcolor[rgb]{ .906,  .902,  .902}\textcolor[rgb]{ 0,  0,  1}{\textbf{(-0.96)}} }
          & \cellcolor[rgb]{ .906,  .902,  .902}\textbf{46.90} \scriptsize{ \textcolor[rgb]{ 0,  0,  1}{(-1.85)} }
          & 34.07 \scriptsize{ \textcolor[rgb]{ 0,  0,  1}{(-0.59)} }
          & 18.30 \scriptsize{ \textcolor[rgb]{ 1,  0,  0}{(+0.09)} }
          \\
    \bottomrule
    \end{tabular}%
    }

\label{over ranges}
\end{table*}%
In Table.\ref{across different categories},
we show the experimental results on the nuScenes val/test set with the state-of-the-arts. Without any test-time augmentation nor model ensemble, our SemanticBEVFusion achieves the superior performance of 70.9$\%$ in mAP and 73.0$\%$ in NDS. 
Our approach outperforms the SOTA LiDAR method, TransFusion-L\cite{bai2022transfusion} by +5.4$\%$ in mAP and +2.8$\%$ in NDS.
Compared with BEVFusion\cite{liang2022bevfusion}, we achieve +1.7$\%$ higher mAP and +1.2$\%$ higher NDS on the test set. Compared with BEVFusion \cite{liu2022bevfusion}, we achieve +0.7$\%$ higher in terms of mAP while reducing the number of pseudo points by average 84.36$\%$ of the original amount for boosted efficiency.
Our approach outperforms other competing SOTA fusing models with evident increment in mAP over most categories.
The performance gain should be ascribed to our 
proposed semantic fusion on camera BEV and LiDAR BEV features. 

\noindent{\textbf{Different Ranges and True Positive Metrics.}}
To comprehensively demonstrate the performance of our approach, we split the objects in whole (0-54m) into near (0-18 m), mid (18-36m), and far (36-54m). In Table.\ref{over ranges}, we compare our approach with the LiDAR-only baseline\cite{liu2022bevfusion, bai2022transfusion} and  BEVFusion \cite{liu2022bevfusion} in terms of mAP, NDS and true Positive metrics including Average Translation Error (ATE), Average Scale Error (ASE), Average Orientation Error (AOE), Average Velocity Error (AVE) and Average Attribute Error (AAE)\cite{nuscenes2019}. Higher AP, NDS and lower ATE, ASE, AOE, AVE, AAE indicate better performance, marked in blue and highlighted. Our approach achieves superior performance in all ranges in terms of mAP and NDS. Especially, in the far range, our approach outperforms BEVFusion \cite{liu2022bevfusion} by 1.53$\%$ in mAP and 1.04$\%$ in NDS, also by 7.30$\%$ in mAP and 4.19$\%$ in NDS compared with LiDAR baseline\cite{liu2022bevfusion, bai2022transfusion}, benefiting from the cross-modal semantic interaction for combining accurate locations with semantic results. 
The semantic information can help with the selection of possible object proposals with approximate locations, better classification regression, and prior category-related information like scale. 
This assumption is confirmed by the unanimous improvement of mATE and mASE in all ranges.
For orientation estimation, camera and LiDAR are complementary over different ranges. LiDAR points are dense enough in near range for accurate orientation. Camera features are relatively denser in the far range to retrieve the observation angle with the ray direction for orientation estimation. Therefore, the fusion of camera geometric information is effective for far-away objects when LiDAR points are too sparse to correctly regress orientation.

\noindent{\textbf{Weather and Lighting. }}
We present the performance of our approach under different weather and lighting conditions in Table.\ref{weather}. We follow the nuScenes descriptor to split night and daytime, rainy and sunny. Our SemanticBEVFusion outperforms BEVFusion \cite{liu2022bevfusion} by 1.09$\%$ in mAP and 0.56$\%$ in NDS under daytime,  1.34$\%$ in mAP and 0.94$\%$ in NDS under rainy conditions, and  0.68$\%$ in mAP and 0.51$\%$ in NDS under sunny conditions. Rainy condition degrades LiDAR points and damages the detection performance of LiDAR-only method\cite{liu2022bevfusion, bai2022transfusion}. Our approach significantly outperforms the LiDAR-only model\cite{liu2022bevfusion, bai2022transfusion} by a large margin, +8.93$\%$ mAP under challenging rainy conditions. We observe  performance degradation of our approach under night conditions. The bad illumination condition during nighttime severely degrades the image quality, as well as the segmentation results. Still, we outperform the LiDAR-only\cite{liu2022bevfusion, bai2022transfusion} method by 3.25$\%$ in mAP and 1.69$\%$ in NDS respectively.
We believe that our approach at night can be boosted with the replacement of stronger 2D backbones as we already qualitatively observe the performance increment by switching  current CenterNet2 by a transformer-based image segmentation model \cite{cheng2021maskformer}. We will release the results afterward due to limited time.

\begin{table*}[t!]
  \centering
  \caption{\textbf{Results under different weather and lighting conditions.} Our approach outperforms BEVFusion\cite{liu2022bevfusion} and the LiDAR-only method\cite{liu2022bevfusion, bai2022transfusion} on daytime, sunny and rainy conditions.}
  \resizebox{\textwidth}{!}{
    \begin{tabular}{l|cc|cc|cc|cc}
    \toprule
    \multirow{2}[2]{*}{Method} 
    & \multicolumn{2}{c|}{Night}    
    & \multicolumn{2}{c|}{Daytime}  
    & \multicolumn{2}{c|}{Rainy}    
    & \multicolumn{2}{c}{Sunny} \\
          
      & mAP 
      & NDS
      & mAP 
      & NDS 
      & mAP 
      & NDS
      & mAP 
      & NDS 
      \\
          
    \midrule
    
    LiDAR-only\cite{liu2022bevfusion, bai2022transfusion}
    & 35.35
    & 42.33
    & 64.41
    & 69.41
    & 62.36
    & 68.94
    & 64.35
    & 69.17
    \\
    
    BEVFusion \cite{liu2022bevfusion}
    & \cellcolor[rgb]{ .906,  .902,  .902}\textbf{43.20} \scriptsize{ \cellcolor[rgb]{ .906,  .902,  .902}\textcolor[rgb]{ 0,  0,  1}{\textbf{(+7.85)}} }
    & \cellcolor[rgb]{ .906,  .902,  .902}\textbf{46.17} \scriptsize{ \cellcolor[rgb]{ .906,  .902,  .902}\textcolor[rgb]{ 0,  0,  1}{\textbf{(+3.84)}} }
    & 68.51 \scriptsize{ \textcolor[rgb]{ 0,  0,  1}{(+4.10)} }
    & 71.58 \scriptsize{ \textcolor[rgb]{ 0,  0,  1}{(+2.17)} }
    & 69.95 \scriptsize{ \textcolor[rgb]{ 0,  0,  1}{(+7.59)} }
    & 72.45 \scriptsize{ \textcolor[rgb]{ 0,  0,  1}{(+3.51)} }
    & 68.29 \scriptsize{ \textcolor[rgb]{ 0,  0,  1}{(+3.94)} }
    & 71.25 \scriptsize{ \textcolor[rgb]{ 0,  0,  1}{(+2.08)} }
    \\

    \textbf{SemanticBEV} 
    & 38.60 \scriptsize{ \textcolor[rgb]{ 0,  0,  1}{(+3.25)} }
    & 44.02 \scriptsize{ \textcolor[rgb]{ 0,  0,  1}{(+1.69)} }
    & \cellcolor[rgb]{ .906,  .902,  .902}\textbf{69.60} \scriptsize{ \cellcolor[rgb]{ .906,  .902,  .902}\textcolor[rgb]{ 0,  0,  1}{\textbf{(+5.19)}} }
    & \cellcolor[rgb]{ .906,  .902,  .902}\textbf{72.14} \scriptsize{ \cellcolor[rgb]{ .906,  .902,  .902}\textcolor[rgb]{ 0,  0,  1}{\textbf{(+2.73)}} }
    & \cellcolor[rgb]{ .906,  .902,  .902}\textbf{71.29} \scriptsize{ \cellcolor[rgb]{ .906,  .902,  .902}\textcolor[rgb]{ 0,  0,  1}{\textbf{(+8.93)}} }
    & \cellcolor[rgb]{ .906,  .902,  .902}\textbf{73.39} \scriptsize{ \cellcolor[rgb]{ .906,  .902,  .902}\textcolor[rgb]{ 0,  0,  1}{\textbf{(+4.45)}} }
    & \cellcolor[rgb]{ .906,  .902,  .902}\textbf{68.97} \scriptsize{ \cellcolor[rgb]{ .906,  .902,  .902}\textcolor[rgb]{ 0,  0,  1}{\textbf{(+4.62)}} }
    & \cellcolor[rgb]{ .906,  .902,  .902}\textbf{71.76} \scriptsize{ \cellcolor[rgb]{ .906,  .902,  .902}\textcolor[rgb]{ 0,  0,  1}{\textbf{(+2.59)}} }\\
    \bottomrule
    \end{tabular}%
    }
  \label{weather}%
\end{table*}%

\subsection{Ablation Studies}
\label{sec:ablation}
\noindent{\textbf{Semantic View Transformation.}}
As in section 3.1, we decompose the semantic view transformation to mapping, depth attention \cite{philion2020lift, liu2022bevfusion}, and semantic masking. The mAP and NDS are improved by 3.42$\%$ and 1.90$\%$ when the mapping is conducted, while a  marginal increase (+0.69$\%$ mAP, +0.42$\%$ NDS) is observed after the depth attention, as shown in Table.\ref{view transformation}. This indicates the dominating contribution of scattering semantic features to BEV during the mapping mechanism. 
Previous BEVFusion methods \cite{liu2022bevfusion, liang2022bevfusion} claim that the view transformation relies on the accurate depth distribution along the ray direction so the spatially aligned camera features and LiDAR features are matched and fused. We think that the mapping process is the core functional part for densely placing the semantic camera features in approximate BEV locations for nearby LiDAR features to fetch during fusion. 

\noindent{\textbf{Fusion Module.}}
In Table.\ref{component analysis} (row \emph{c} and \emph{d}), ablation experiments on two different fusion modules are shown: a convolutional fuser and an additive fuser. Convolutional fuser with the broader receptive field may compensate for the inaccurate depth prediction as claimed in BEVFusion\cite{liu2022bevfusion}. However, no obvious improvement is seen in our SemanticBEVFusion. The results confirm the importance of semantic view mapping, compared with depth attention.

\noindent{\textbf{Camera Components.}}
To validate the design of our camera stream, we conduct ablation experiments to quantify the contribution of each component proposed in our model. In Table.\ref{component analysis} row \emph{a} and row \emph{b},
the sole introduction of semantic points surpasses LiDAR-only\cite{liu2022bevfusion, bai2022transfusion} method by 4.31$\%$ in mAP and 2.39$\%$ in NDS. 
In Table.\ref{component analysis} row \emph{b} and row \emph{d}, adding semantic embedded camera features further boosts the detection performance by 0.92$\%$ in mAP and 0.38$\%$ in NDS.
In Table.\ref{component analysis} row \emph{e} and \emph{f}, semantic masking achieves comparable detection performance while reducing the pseudo points by an average of 84.36$\%$. This confirms the effectiveness of semantic features with masking which filters out the irrelevant background features. The marginal gap can be closed by a more accurate instance segmentation model.

\begin{table*}[htbp]
   \centering
  \caption{\textbf{Component analysis of our model on nuScenes validation set.} The joint introduction of both semantic camera features and geometric camera features can significantly boost the detection performance with an increase of 5.23$\%$ in mAP and 2.77$\%$ in NDS.}
  \resizebox{\textwidth}{!}{
    \begin{tabular}{c|cc|ccc|cc|cc}
    \toprule
    \multirow{2}[2]{*}{ID} & \multicolumn{2}{c|}{LiDAR Stream} & \multicolumn{3}{c|}{Camera Stream} & \multicolumn{2}{c|}{Fuser} & \multicolumn{2}{c}{Metrics} \\
          & XYZI & Semantic &  Semantic Embedded Features & Semantic Masking & Depth Attention Masking & Conv  & Add & mAP   & NDS \\
    \midrule
    a)    & \Checkmark  &       &       &       &       &       &  & 64.24 & 69.19  \\
    b)    & \Checkmark & \Checkmark  &       &       &       &       &  & 68.55  & 71.58  \\
    
    \midrule
    c)    & \Checkmark & \Checkmark  & \Checkmark  &       &       &       & \Checkmark  & 69.39 & 71.91  \\
    d)    & \Checkmark & \Checkmark  &  \Checkmark  &       &       & \Checkmark  &       & 69.47 & 71.96  \\

    \midrule
    e)    & \Checkmark  & \Checkmark   & \Checkmark  &       & \Checkmark  & \Checkmark  &       & 68.62 & 71.37  \\
    \textbf{f)} & \Checkmark & \textbf{\Checkmark} & \textbf{\Checkmark} & \textbf{\Checkmark} &       & \textbf{\Checkmark} &       & \textbf{69.24} & \textbf{71.85}  \\
    \bottomrule
    \end{tabular}%
    }

  \label{component analysis}%
\end{table*}%

\noindent{\textbf{Different Query Number at Inference.}}
The non-parametric initialization for the queries allows the modification of query number during inference\cite{bai2022transfusion}. Our approach is trained with 200 queries and the performance of different queries number during inference is evaluated and listed in Table.\ref{query number}. The performance increases and tends to be stable when the query number exceeds 200. We use the results of 300 queries for slightly better performance.

\begin{minipage}{\textwidth}
\begin{minipage}[t]{0.58\textwidth}
\makeatletter\def\@captype{table}
\centering
  \caption{Ablating camera-BEV transformation mechanisms.}
  \resizebox{0.55\textwidth}{!}{
  
    \begin{tabular}{ccccc}
    \toprule
    Mapping & Depth Attention  & mAP  & NDS \\
    \midrule
          &       & 64.24 & 69.19  \\
    \Checkmark     &        & 67.66 & 71.09  \\
    \Checkmark     & \Checkmark     & 68.35  & 71.51  \\

    \bottomrule
    \end{tabular}%
    }
    
\label{view transformation}%
\end{minipage}
\begin{minipage}[t]{0.38\textwidth}
\makeatletter\def\@captype{table}
\centering
\caption{Ablating the query number. }
\resizebox{\textwidth}{!}{

    \begin{tabular}{ccccc}
    \toprule
     topN  & 100   & 200   & 300   & 400 \\
    \midrule
     mAP   & 68.38 & 69.45 & 69.61 & 69.64 \\
    NDS   & 71.44 & 71.95 & 71.97 & 72.01 \\
    \bottomrule
    \end{tabular}%
    }
\label{query number}%
\end{minipage}
\end{minipage}

\subsection{Future work}
\label{sec:future work}
The current model still has unsatisfying performance under night and over-exposure conditions. We believe a transformer-based backbone \cite{cheng2021maskformer} can enhance robustness to illuminations and further boost the detection performance. In addition, the superiority of our approach increases as the range increases. We would like to further investigate performance on far objects which are beyond nuScenes range: 54m since distant object detection is extremely important for L2+ high-way ADAS applications. We also hope that our work will inspire further investigation of semantic and geometric influences during multi-modality fusion.

\section{Conclusion}
\label{sec:Conclusion}

In this paper, we rethink the current popular BEVFusion strategies and analyze the semantic and geometric representational characteristics of LiDAR features and camera features on 3D object detection. We present SemanticBEVFusion, which carries on camera and LiDAR semantic and geometric representational strengths in one elegant BEV representation and boosts 3D detection performance. We claim and prove the necessity of semantic fusion in view transformation. Our method achieves state-of-the-art performance on the large-scale nuScenes 3D object detection leaderboard.

\end{document}